%% file: acl2018.tex
\documentclass[11pt,a4paper]{article}
\usepackage[hyperref]{acl2018}
\usepackage{times}
\usepackage{latexsym}

\usepackage{url}
\include{mlVecMat}

\newcommand{\RN}[1]{%
	\textup{\lowercase\expandafter{\it \romannumeral#1}}%
}

\aclfinalcopy % Uncomment this line for the final submission
\def\aclpaperid{1006} %  Enter the acl Paper ID here

\title{Joint Embedding of Words and Labels for Text Classification}

\author{Guoyin Wang, Chunyuan Li\thanks{~~Corresponding author}, Wenlin Wang, Yizhe Zhang\\ 
	 \textbf{Dinghan Shen, Xinyuan Zhang, Ricardo Henao, Lawrence Carin}  \\
  Duke University \\
  {\tt \{gw60,cl319,ww107,yz196,ds337,xz139,r.henao,lcarin\}@duke.edu} \\}

\date{}

\begin{document}
\maketitle
\begin{abstract}
Word embeddings are effective intermediate representations for capturing semantic regularities between words, when learning the representations of text sequences. We propose to view text classification as a label-word joint embedding problem: each label is embedded in the same space with the word vectors. We introduce an attention framework that measures the compatibility of embeddings between text sequences and labels. The attention is learned on a training set of labeled samples to ensure that, given a text sequence, the relevant words are weighted higher than the irrelevant ones. Our method maintains the interpretability of word embeddings, and enjoys a built-in ability to leverage alternative sources of information, in addition to input text sequences.  Extensive results on the several large text datasets show that the proposed framework outperforms the state-of-the-art methods by a large margin, in terms of both accuracy and speed.

\end{abstract}

\section{Introduction}
Text classification is a fundamental problem in natural language processing (NLP). The task is to annotate a given text sequence with one (or multiple) class label(s) describing its textual content. 
A key intermediate step is the text representation. Traditional methods represent text with hand-crafted features, such as sparse lexical features (\eg $n$-grams)~\cite{wang2012baselines}. Recently, neural models have been employed to learn text representations, including convolutional neural networks (CNNs) \cite{kalchbrenner2014convolutional,zhang2017deconvolutional,shen2017deconvolutional} and recurrent neural networks (RNNs) based on long short-term memory (LSTM)~\cite{hochreiter1997long,wang2018topic}.

To further increase the representation flexibility of such models, attention mechanisms~\cite{bahdanau2015neural} have been introduced as an integral part of models employed for text classification~\cite{yang2016hierarchical}. The attention module is trained to capture the dependencies that make significant contributions to the task, regardless of the distance between the elements in the sequence. It can thus provide complementary information to the distance-aware dependencies modeled by RNN/CNN. The increasing representation power of the attention mechanism comes with increased model complexity.

% The recent self-attention produces context-aware representation by applying attention to each pair of tokens from the input sequence, which characterizes the intra semantic correlation among the words of different positions.

Alternatively, several recent studies show that the success of deep learning on text classification largely depends on the effectiveness of the word embeddings
~\cite{joulin2016bag,wieting2016towards,arora2017simple,shen2018on}.
Particularly, \citet{shen2018on} quantitatively show that the word-embeddings-based text classification tasks can have the similar level of difficulty regardless of the employed models, using the concept of intrinsic dimension~\cite{li_id_2018_ICLR}. Thus, simple models are preferred.  As the basic building blocks in neural-based NLP, word embeddings capture the similarities/regularities between words~\cite{mikolov2013distributed,pennington2014glove}. This idea has been extended to compute embeddings that capture the semantics of word sequences (\eg phrases, sentences, paragraphs and documents)~\cite{le2014distributed,kiros2015skip}. These representations are built upon various types of compositions of word vectors, ranging from simple averaging to sophisticated architectures. Further, they suggest that simple models are efficient and interpretable, and have the potential to outperform sophisticated deep neural models. 

It is therefore desirable to leverage the best of both lines of works: learning text representations to capture the dependencies that make significant contributions to the task, while maintaining low computational cost. For the task of text classification, labels play a central role of the final performance. A natural question to ask is how we can directly use label information in constructing the text-sequence representations.

\subsection{Our Contribution}
Our primary contribution is therefore to propose such a
solution by making use of the label embedding framework, and propose the  
{\em Label-Embedding Attentive Model} (LEAM) to improve text classification.
While there is an abundant literature in the NLP community on word embeddings (how to describe a word) for text representations, much less work has been devoted in comparison to label embeddings (how to describe a class). The proposed LEAM is implemented by jointly embedding the word and label in the same latent space, and the text representations are constructed directly using the text-label compatibility.

Our label embedding framework has the following salutary properties:
$(\RN{1})$ Label-attentive text representation is informative for the downstream classification task, as it directly learns from a shared joint space, whereas traditional methods proceed in multiple steps by solving intermediate problems.
$(\RN{2})$  The LEAM learning procedure only involves a series of basic algebraic operations, and hence it retains the interpretability of simple models, especially when the label description is available.
$(\RN{3})$ Our attention mechanism (derived from the text-label compatibility) has fewer parameters and less computation than related methods, and thus is much cheaper in both training and testing, compared with sophisticated deep attention models.
$(\RN{4})$  We perform extensive experiments on several text-classification tasks, demonstrating the effectiveness of our label-embedding attentive model, providing state-of-the-art results on benchmark datasets.
$(\RN{5})$ We further apply LEAM to predict the medical codes from clinical text. As an interesting by-product, our attentive model can highlight the informative key words for prediction, which in practice can reduce a doctor's burden on reading clinical notes.

\section{Related Work}
Label embedding has been shown to be effective in various domains and tasks.
In computer vision, there has been a vast amount of research on leveraging label embeddings for image classification~\citep{akata2016label}, multimodal learning between images and text~\citep{frome2013devise,kiros2014unifying}, and text recognition in images~\cite{rodriguez2013label}. It is particularly successful on the task of zero-shot learning~\cite{palatucci2009zero,yogatama2015embedding,ma2016label}, where the label correlation captured in the embedding space can improve the prediction when some classes are unseen.  
In NLP, labels embedding for text classification has been studied in the context of heterogeneous networks in~\citep{tang2015pte} and multitask learning in~\citep{zhang2017multi}, respectively.
To the authors' knowledge, there is little research on investigating the effectiveness of label embeddings to design efficient attention models, and how to joint embedding of words and labels to make full use of label information for text classification has not been studied previously, representing a contribution of this paper.

For text representation, the currently best-performing models usually consist of an encoder and a decoder connected through an attention mechanism~\cite{vaswani2017attention,bahdanau2015neural}, with successful applications to sentiment classification~\cite{zhou2016attention}, sentence pair modeling~\cite{yin2016abcnn} and sentence summarization~\cite{rush2015neural}. Based on this success, more advanced attention models have been developed, including hierarchical attention networks~\cite{yang2016hierarchical}, attention over attention~\cite{cui2016attention}, and multi-step attention~\cite{gehring2017convolutional}.
The idea of attention is motivated by the observation that different words in the same context are differentially informative, and the same word may be differentially important in a different context. The realization of ``context'' varies in different applications and model architectures. Typically, the context is chosen as the target task, and the attention is computed over the hidden layers of a CNN/RNN. 
Our attention model is directly built in the joint embedding space of words and labels, and the context is specified by the label embedding.

Several recent works~\cite{vaswani2017attention,shen2018disan,shen2018bi} have demonstrated that simple attention architectures can alone achieve state-of-the-art performance with less computational time, dispensing with recurrence and convolutions entirely. Our work is in the same direction, sharing the similar spirit of retaining model simplicity and interpretability. The major difference is that the aforementioned work focused on self attention, which applies attention to each pair of word tokens from the text sequences. In this paper, we investigate the attention between words and labels, which is more directly related to the target task. Furthermore, the proposed LEAM has much less model parameters. 
% Since label embeddings are the representative points of the text sequence embeddings, one may view LEAM as an efficient approximation to the ``self-attention''.

\section{Preliminaries}
%use $\odot$ for element-wise multiplication, $\oslash$ for element-wise division and $\oplus$ for concatenation
%
Throughout this paper, we denote vectors as bold, lower-case letters, and matrices as bold, upper-case
letters. We use $\oslash$ for element-wise division when applied to vectors or matrices. We use $\circ$ for function composition, and $\Delta^p$ for the set of one hot vectors in dimension $p$.

Given a training set $\Scal  = \{ (\Xmat_n, \yv_n) \}_{n=1}^N$ of pair-wise data,
where $\Xmat \in \Xcal$  is the text sequence, and $\yv \in \Ycal$ is its corresponding label. Specifically, $\yv$ is a one hot vector in single-label problem and a binary vector in multi-label problem, as defined later in Section 4.1.
Our goal for text classification is to learn a function $f: \Xcal  \mapsto \Ycal$ by minimizing an empirical risk of the form:
	\vspace{-2mm}
\beq
\min_{f \in \Fcal} \frac{1}{N} \sum_{n=1}^{N} 
\delta (\yv_n, f(\Xmat_n) )
	\vspace{-2mm}
\label{eq:objective}
\eeq
where $\delta: \Ycal \times \Ycal \mapsto \R $  measures the loss incurred from
predicting $f(\Xmat)$ when the true label is $\yv$, where $f$ belongs to the functional space $\Fcal$. 
In the evaluation stage, we shall use the $0/1$ loss as a target loss: $\delta (\yv, \zv) = 0$ if $\yv = \zv$, and $1$ otherwise. 
In the training stage, we consider surrogate losses commonly used for structured prediction in different problem setups (see Section~\ref{sec:model} for details on the surrogate losses used
in this paper).

\begin{figure}[t!] \centering
	\vspace{-0mm}
	\begin{tabular}{c}		
		\includegraphics[width=7.40cm]{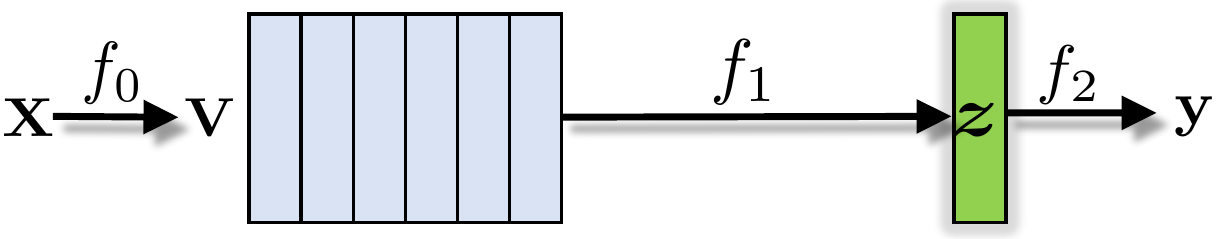} \\
		(a) Traditional method  \vspace{2mm}  \\
		\includegraphics[width=7.40cm]{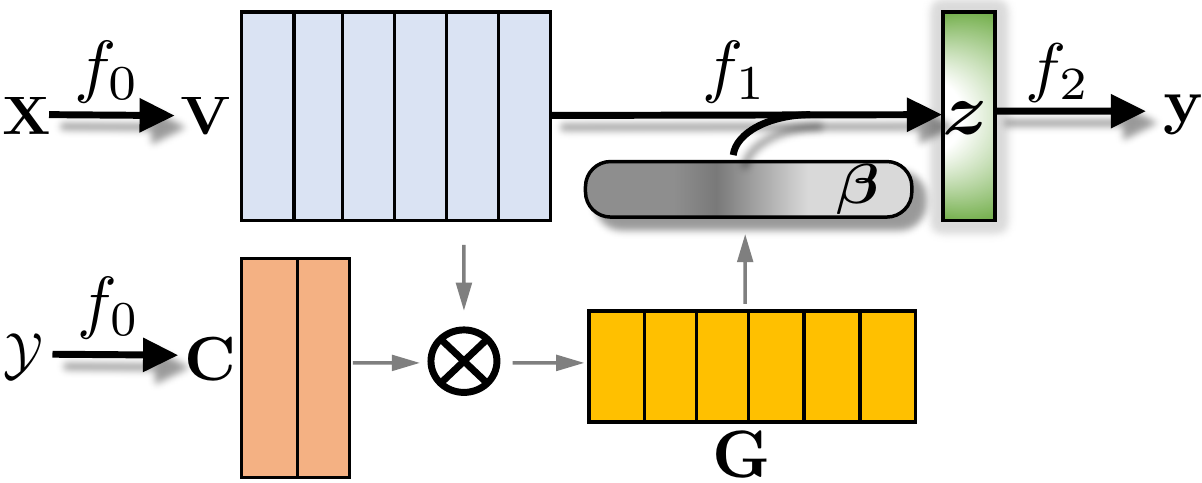}   \\
		(b) Proposed joint embedding method \hspace{-0mm}
	\end{tabular}
	\vspace{-0mm}
	\caption{Illustration of different schemes for document representations $\zv$. 
		(a) Much work in NLP has been devoted to directly aggregating word embedding $\Vmat$ for $\zv$. (b) We focus on learning label embedding $\Cmat$ (how to embed class labels in a Euclidean space), and leveraging the ``compatibility'' $\Gmat$ between embedded words and labels to derive the attention score $\betav$ for improved $\zv$.
		Note that $\otimes$ denotes the cosine similarity between $\Cmat$ and $\Vmat$. In this figure, there are K=2 classes. }
	\vspace{-0mm}
	\label{fig:model_arch}
\end{figure}

More specifically, an input sequence $\Xmat$ of length $L$ is composed of word tokens: $\Xmat = \{\xv_1, \cdots, \xv_L\}$. Each token $\xv_{l}$ is a one hot vector in the space $\Delta^{D}$, where $D$ is the dictionary size. Performing learning in $\Delta^{D}$ is computationally expensive and difficult.
An elegant framework in NLP, initially proposed in~\cite{mikolov2013distributed,le2014distributed,pennington2014glove,kiros2015skip}, allows to concisely perform learning by mapping the words into an embedding space. The framework relies on so called {\em word embedding}: $\Delta^D \mapsto \R^P$, where $P$ is the dimensionality of the embedding space.
Therefore, the text sequence $\Xmat$ is represented via the respective word embedding for each token: $\Vmat = \{\vv_1, \cdots, \vv_L\}$, where $\vv_l \in \R^P$. A typical text classification method proceeds in three steps, end-to-end, by considering a function decomposition $f=f_0 \circ  f_1 \circ  f_2$ as shown in Figure~\ref{fig:model_arch}(a):
\begin{itemize}
	\item $f_0: \Xmat \mapsto \Vmat$, the text sequence is represented as its word-embedding form $\Vmat$, which is a matrix of $P \times L$. 
		\vspace{-2mm}
	\item $f_1: \Vmat \mapsto \zv$,  a compositional function $f_1$ aggregates word embeddings into a fixed-length vector representation $\zv$.
		\vspace{-2mm}
	\item $f_2: \zv \mapsto \yv$, a classifier $f_2$ annotates the text representation $\zv$ with a label.
\end{itemize}
A vast amount of work has been devoted to devising the proper functions $f_0$ and $f_1$, \ie how to represent a word or a word sequence, respectively. The success of NLP largely depends on the effectiveness of word embeddings in $f_0$~\cite{bengio2003neural,collobert2008unified,mikolov2013distributed,pennington2014glove}. They are often pre-trained offline on large corpus, then refined jointly via $f_1$ and $ f_2$ for task-specific representations. 
Furthermore, the design of $f_1$ can be broadly cast into two categories. The popular deep learning models consider the mapping as a ``black box,'' and have employed sophisticated CNN/RNN architectures to achieve state-of-the-art performance~\cite{zhang2015character,yang2016hierarchical}. On the contrary, recent studies show that simple manipulation of the word embeddings, \eg mean or max-pooling, can also provide surprisingly excellent performance~\cite{joulin2016bag,wieting2016towards,arora2017simple,shen2018on}. Nevertheless, these methods only leverage the information from the input text sequence.

\section{Label-Embedding Attentive Model}

\subsection{Model}~\label{sec:model}
By examining the three steps in the traditional pipeline of text classification, we note that the use of label information only occurs in the last step, when learning $f_2$, and its impact on learning the representations of words in $f_0$ or word sequences in $f_1$ is ignored or indirect. Hence, we propose a new pipeline by incorporating label information in every step, as shown in Figure~\ref{fig:model_arch}(b):
\begin{itemize}
	\item $f_0$: Besides embedding words, we also embed all the labels in the same space, which act as the ``anchor points'' of the classes to influence the refinement of word embeddings.
	\item $f_1$: The compositional function aggregates word embeddings into $\zv$, weighted by the compatibility between labels and words.
	\item $f_2$: The learning of $f_2$ remains the same, as it directly interacts with labels.
\end{itemize}
Under the proposed label embedding framework, we specifically describe a label-embedding attentive model.

\paragraph{Joint Embeddings of Words and Labels}
% Instead of considering $f_1$ as a one-directional mapping from word embeddings $\Vmat$ to document representation $\zv$, it is desirable to incorporate label information in constructing $\zv$. 
%
We propose to embed both the words and the labels into a joint space \ie $\Delta^D \mapsto \R^P$ and  $\Ycal  \mapsto \R^P$. The label embeddings are $\Cmat = [\cv_1, \cdots, \cv_K ]$, where $K$ is the number of classes.

A simple way to measure the compatibility of label-word pairs is via the cosine similarity
\beq
\Gmat = (\Cmat^{\top} \Vmat)  \oslash  \hat{\Gmat},
\label{eq:label_word_linear}
\eeq	
where $\hat{\Gmat}$ is the normalization matrix of size $K \times L$, with each element obtained as the multiplication of $\ell_2$ norms of the $c$-th label embedding and $l$-th word embedding: $\hat{g}_{kl} =\| \cv_k \| \| \vv_l \|$.
% The compatibility matrix $\Gmat$ is of size $K \times L$. To have normalized measure, we can define the cosine similarity  $\Gmat=\Vmat^{nT} \Lmat^{n}$, where each entry $\Gmat_{i,j}$ represents the cosine similarity between word $i$ and class $j$. 

% Also words do not contribute equally to the representation of the sentence meaning, especially for classification tasks. Similarly as Hierarchical Attention \cite{yang2016hierarchical} and Block Self-Attention \cite{shen2018bi}, we want to evaluate the importance of a word based on itself and its neighbors. Hence, we also use attention mechanism to extract words that are important to the meaning of the sentence/label and aggregate the representation of those words to generate a sentence vector. Specifically,

To further capture the relative spatial information among consecutive words (\ie phrases\footnote{We call it ``phrase'' for convenience; it could be any longer word sequence such as a sentence and paragraph \etc when a larger window size $r$ is considered.}) and introduce non-linearity in the compatibility measure, we consider a generalization of~\eqref{eq:label_word_linear}. Specifically, for a text phase of length $2r+1$ centered at $l$, the local matrix block $\Gmat_{l-r:l+r}$ in $\Gmat$ measures the label-to-token compatibility for the ``label-phrase'' pairs. To learn a higher-level compatibility stigmatization $\uv_{l}$ between the $l$-th phrase and all labels, we have:
\beq{
	\uv_{l} = \mbox{ReLU}(  \Gmat_{l-r:l+r}  \Wmat_1 + \bv_1),
	\label{eq:label_word_nonlinear}
}\eeq
where $\Wmat_1 \in \R^{ 2r+1 }$ and $\bv_1 \in \R^{ K }$ are parameters to be learned, and $\uv_l\in\mathbb{R}^K$.
The largest compatibility value of the $l$-th phrase \wrt the labels is collected:
\beq{
	m_l =  \mbox{max-pooling}(\uv_{l}).
}\eeq
Together, $\mv$ is a vector of length $L$. 
% , each element representing the maximum compatibility value among the labels. 
%
The compatibility/attention score for the entire text sequence is:
\beq{
	\betav = \mbox{SoftMax}(\mv),
}\eeq
where the $l$-th element of SoftMax is $\beta_l = \frac{\exp(m_l)}{\sum_{l'=1}^L \exp(m_{l'}) }$.

The text sequence representation can be simply obtained via averaging the word embeddings, weighted by label-based attention score:
\beq{
	\zv = \sum_l \beta_l  \vv_l.
	\label{eq:text_vector}
}\eeq

% That is, we first update the word $i$ correlation with class $j$ considering its adjacency words $i-r: i+r$ by one-layer MLP. Then we extract most significant correlation score as the word importance measure. By passing through a softmax function, we obtain the normalized importance weight $\betav$. Finally, we compute the sentence vector $\zv$  as a weighted sum of the full word embedding based on the importance weight.

\paragraph{Relation to Predictive Text Embeddings} 
Predictive Text Embeddings (PTE)~\citep{tang2015pte} is the first method to leverage label embeddings to improve the learned word embeddings. We discuss three major differences between PTE and our LEAM: 
$(\RN{1})$~The general settings are different. PTE casts the text representation through heterogeneous networks, while we consider text representation through an attention model.
$(\RN{2})$~In PTE, the text representation $\zv$ is the averaging of word embeddings. In LEAM, $\zv$ is weighted averaging of word embeddings through the proposed label-attentive score in \eqref{eq:text_vector}.
$(\RN{3})$~PTE only considers the linear interaction between individual words and labels. LEAM greatly improves the performance by considering nonlinear interaction between phrase and labels. 
Specifically, we note that the text embedding in PTE is similar with a very special case of LEAM, when our window size $r=1$ and attention score $\betav$ is uniform. As shown later in Figure~\ref{fig:leam_details}(c) of the experimental results, LEAM can be significantly better than the PTE variant.

\paragraph{Training Objective}
The proposed joint embedding framework is applicable to various text classification tasks. We consider two setups in this paper. For a learned text sequence representation $\zv = f_1 \circ f_0(\Xmat)$, we jointly optimize $f=f_0 \circ f_1 \circ f_2$ over $\Fcal$, where $f_2$ is defined according to the specific tasks:
\begin{itemize}
	\item {\em Single-label problem}: categorizes each text instance to precisely one of $K$ classes, $\yv \in \Delta^K$
	\beq
	\min_{f \in \Fcal} \frac{1}{N} \sum_{n=1}^{N}
	\mbox{CE} (\yv_{n}, f_2(\zv_{n}) ),
	\label{eq:objective_multiclass}
	\eeq
	where $\mbox{CE}(\cdot, \cdot)$ is the cross entropy between two probability vectors, 
	and $f_2(\zv_n)= \mbox{SoftMax }( \zv_n^{\prime} ) $, with $\zv_n^{\prime} = \Wmat_2 \zv_{n} + \bv_2$ 
	and $ \Wmat_2 \in \R^{K \times P} ,  \bv_2 \in \R^K$ are trainable parameters.
	\item {\em Multi-label problem}: categorizes each text instance to a set of $K$ target labels $\{\yv_k \in \Delta^2 | k=1,\cdots, K\}$; there is no constraint on how many of the classes the instance can be assigned to, and
	\beq
	\min_{f \in \Fcal} \frac{1}{NK} \sum_{n=1}^{N}  \sum_{k=1}^{K} 
	\mbox{CE} (\yv_{nk}, f_{2}(\zv_{nk} ),
	\label{eq:objective_multilabel}
	\eeq
	where $f_{2}(\zv_{nk}) = \frac{1}{1+\exp( \zv_{nk}^{\prime}  ) }$, and $\zv_{nk}^{\prime}$ is the $k$th column of $ \zv_{n}^{\prime}$.
\end{itemize}
To summarize, the model parameters 
$\thetav = \{\Vmat, \Cmat, \Wmat_1, \bv_1, \Wmat_2, \bv_2\}$. They are trained end-to-end during learning. $\{ \Wmat_1, \bv_1\}$ and $\{\Wmat_2, \bv_2\}$ are weights in $f_1$ and $f_2$, respectively, which are treated as standard neural networks. For the joint embeddings $\{\Vmat, \Cmat\}$ in $f_0$, the pre-trained word embeddings are used as initialization if available.

\subsection{Learning \& Testing with LEAM}

\paragraph{Learning and Regularization}
The quality of the jointly learned embeddings are key to the model performance and interpretability. Ideally, we hope that each label embedding acts as the ``anchor'' points for each classes: closer to the word/sequence representations that are in the same classes, while farther from those in different classes. 
To best achieve this property, we consider to regularize the learned label embeddings $\cv_k$ to be on its corresponding manifold. This is imposed by the fact $\cv_k$ should be easily classified as the correct label $\yv_k$:
\beq
\min_{f \in \Fcal} \frac{1}{K} \sum_{n=1}^{K}
\mbox{CE} (\yv_{k}, f_2(\cv_{k}) ),
\label{eq:objective_label}
\eeq
where $f_2$ is specficied according to the problem in either \eqref{eq:objective_multiclass} or \eqref{eq:objective_multilabel}.
This regularization is used as a penalty in the main training objective in \eqref{eq:objective_multiclass} or \eqref{eq:objective_multilabel}, and the default weighting hyperparameter is set as 1. It will lead to meaningful interpretability of learned label embeddings as shown in the experiments.

%To best achieve this property, we consider a coupled initialization scheme in the joint embedding space. 
%%

Interestingly in text classification, the class itself is often described as a set of $E$ words $\{\ev_i, i=1, \cdots, E\}$. 
These words are considered as the most representative description of each class, and highly distinguishing between different classes.
For example, the $\mathtt{Yahoo!~Answers~Topic}$ dataset~\cite{zhang2015character} contains ten classes, most of which have two words to precisely describe its class-specific features, such as ``Computers \& Internet'', ``Business \& Finance'' as well as ``Politics \& Government'' \etc 
We consider to use each label's corresponding pre-trained word embeddings as the initialization of the label embeddings.  
% By leveraging the semantic patterns in the generic word embeddings, each label is more likely to initialize closer to the manifold of its own class. 
% This technique will lead to faster learning convergence and better performance as shown in the experiments. 
%
For the datasets without representative class descriptions, one may initialize the label embeddings as random samples drawn from a standard Gaussian distribution.

% \subsection{Testing with the Joint Embeddings}
%
\paragraph{Testing}
Both the learned word and label embeddings are available in the testing stage.
We clarify that the label embeddings $\Cmat$ of all class candidates $\Ycal$ are considered as the input in the testing stage; one should distinguish this from the use of groundtruth label $\yv$ in prediction.
For a text sequence $\Xmat$, one may feed it through the proposed pipeline for prediction: 
$(\RN{1})$ $f_1$: harvesting the word embeddings $\Vmat$, 
$(\RN{2})$ $f_2$: $\Vmat$ interacts with $\Cmat$ to obtain $\Gmat$, pooled as $\betav$, which further attends $\Vmat$ to derive $\zv$, and
$(\RN{3})$ $f_3$: assigning labels based on the tasks. To speed up testing, one may store $\Gmat$ offline, and avoid its online computational cost.
%

%Alternatively, we suggest a faster testing method by re-writing~\eqref{eq:text_vector}:
%%
%\beq{
%	\zv = \sum_l  \hat{\vv}_l, ~\text{where}~\hat{\vv}_l = \beta_l  \vv_l,
%	\label{eq:text_vector2}
%}\eeq 
%%
%where $\hat{\vv}$ is the modulated word embeddings via the label-based attention score. In practice, one may store $\hat{\vv}$ rather than $\vv$ after training.
%Therefore, the test sequence representation is simply the averaging over $\hat{\Vmat}$, yielding the same low testing cost with simple models~ \citep{shen2018on} that average over $\Vmat$.

\subsection{Model Complexity}
We compare CNN, LSTM, Simple Word Embeddings-based Models (SWEM)~\cite{shen2018on} and our LEAM \wrt the parameters and computational speed. For the CNN, we assume the same size $m$ for all filters. Specifically, $h$ represents the dimension of the hidden units in the LSTM or the number of filters in the CNN; $R$ denotes the number of blocks in the Bi-BloSAN; $P$ denotes the final sequence representation dimension. Similar to \cite{vaswani2017attention,shen2018on}, we examine the number of compositional parameters, computational complexity and sequential steps of the four methods.  As shown in Table \ref{tab:archit_compare}, both the CNN and LSTM have a large number of compositional parameters. Since $K \ll m,h$, the number of parameters in our models is much smaller than for the CNN and LSTM models. For the computational complexity, our model is almost same order as the most simple SWEM model, and is smaller than the CNN or LSTM by a factor of $mh/K$ or $h/K$.% Additionally, LEAM is highly parallelizable. 
\begin{table}
	\small
	\begin{adjustbox}{scale=0.8,tabular=l|cccc,center}
		\hline
		\textbf{Model}   & Parameters \! & \! Complexity  & \hspace{-5mm} Seq. Operation \\
		\hline
		CNN & $m \cdot h \cdot P $ & $O(m \cdot h  \cdot  L \cdot P )$ & $O(1)$ \\
		LSTM & $4 \cdot h \cdot (h + P)$ & $O(L \cdot h^2 + h \cdot L \cdot P)$ & $O(L)$ \\
		SWEM & 0 & $O(L \cdot P)$ & $O(1)$ \\
		Bi-BloSAN & $7 \! \cdot \! P^2 \! + \! 5 \! \cdot \! P$ \hspace{-5mm}& \hspace{-5mm} $O(P^2 \!  \cdot \! L^2 \! /  \! R \!+\! P^2  \! \cdot \! L \! + \! P^2 \! \cdot \! R^2)$ \hspace{-5mm} & $O(1)$\\
		Our model & $K \cdot P$  & $O(K \cdot L \cdot P)$ & $O(1)$ \\\hline
	\end{adjustbox}
	\caption{Comparisons of CNN, LSTM, SWEM and our model architecture. Columns correspond to the number of compositional parameters, computational complexity and sequential operations}
	\label{tab:archit_compare}
	\vspace{-4mm}
\end{table}

\section{Experimental Results}
\paragraph{Setup} We use 300-dimensional GloVe word embeddings \citet{pennington2014glove} as initialization for word embeddings and label embeddings in our model. Out-Of-Vocabulary (OOV) words are initialized from a uniform distribution with range $[-0.01, 0.01]$. The final classifier is implemented as an MLP layer followed by a sigmoid or softmax function depending on specific task. We train our model's parameters with the Adam Optimizer \citep{kingma2014adam}, with an initial learning rate of $0.001$, and a minibatch size of 100. Dropout regularization \citep{srivastava2014dropout} is employed on the final MLP layer, with dropout rate $0.5$. The model is implemented using Tensorflow and is trained  on GPU Titan X. %

The code to reproduce the experimental results is at~
{\small  \url{https://github.com/guoyinwang/LEAM}}
\subsection{Classification on Benchmark Datasets}
We test our model on the same five standard benchmark datasets as in \citep{zhang2015character}. The summary statistics of the data are shown in Table \ref{tab:datasets}, with content specified below:
\begin{itemize}
	\item $\mathtt{AGNews}$: Topic classification over four categories of Internet news articles~\citep{del2005ranking} composed of titles plus description classified into: World, Entertainment, Sports and Business. 
	\item $\mathtt{Yelp~~Review~~Full}$: The dataset is obtained from the Yelp Dataset Challenge in 2015, the task is sentiment classification of polarity star labels ranging from 1 to 5.
	\item $\mathtt{Yelp~~Review~~Polarity}$: The same set of text reviews from Yelp Dataset Challenge in 2015, except that a coarser sentiment definition is considered: 1 and 2 are negative, and 4 and 5 as positive. 
	\item $\mathtt{DBPedia}$: Ontology classification over fourteen non-overlapping classes picked from DBpedia 2014 (Wikipedia).
	\item 
	$\mathtt{Yahoo!~~Answers~~Topic}$: Topic classification over ten largest main categories from Yahoo! Answers Comprehensive Questions and Answers version 1.0, including question title, question content and best answer. 
\end{itemize}

\begin{table}
	\centering
	\begin{adjustbox}{scale=0.9,tabular=l|ccc,center}
		\hline
		\textbf{Dataset} & \textbf{\# Classes} & \textbf{\# Training} & \textbf{\# Testing} \\
		\hline
		AGNews & 4 & 120k & 7.6k  \\
		Yelp Binary & 2 & 560 k & 38k  \\
		Yelp Full & 5 & 650k & 38k  \\
		DBPedia & 14 & 560k & 70k  \\
		Yahoo & 10 & 1400k & 60k  \\ 
		\hline
	\end{adjustbox}
	\caption{Summary statistics of five datasets, including the number of classes, number of training samples and number of testing samples.}
	\label{tab:datasets}
\end{table}

\begin{table*}[t!]
	\centering
	\begin{tabular}{c c c c c c}
		\hline
		\textbf{Model} & \textbf{Yahoo } & \textbf{DBPedia} & \textbf{AGNews} & \textbf{Yelp P.} & \textbf{Yelp F.}\\
		\hline
		Bag-of-words~\citep{zhang2015character} & 68.90 & 96.60 & 88.80  & 92.20 & 58.00\\
		Small word CNN~\citep{zhang2015character} & 69.98 & 98.15 & 89.13 & 94.46 & 58.59\\
		Large word CNN \citep{zhang2015character} & 70.94 & 98.28 & 91.45 & 95.11 & 59.48\\
		LSTM  \citep{zhang2015character} & 70.84 & 98.55 & 86.06 & 94.74 & 58.17\\
		SA-LSTM (word-level) \citep{dai2015semi} & - & 98.60 & - & - & -\\
		%Deep CNN (9 layer) & 72.40 & 98.65 & 90.83 & 95.12 & 63.27\\
		%Deep CNN (17 layer) & 72.65 & 98.60 & 91.12 & 95.50 & 63.93\\
		Deep CNN (29 layer) \citep{conneau2017very} & 73.43 & 98.71 & 91.27 & \textbf{95.72} & \textbf{64.26}\\
		SWEM \citep{shen2018on} & 73.53 & 98.42 & 92.24 & 93.76 & 61.11\\
		fastText \citep{joulin2016bag} & 72.30 & 98.60 & 92.50 & 95.70 & 63.90 \\ 
		HAN \citep{yang2016hierarchical} & 75.80 & - & -& -& -\\ % \textbf{71.00}
		Bi-BloSAN$^{\diamond}$ \citep{shen2018bi} & 76.28 & 98.77 & \textbf{93.32} & 94.56 & 62.13 \\
		\hline
		LEAM & \textbf{77.42} & \textbf{99.02} &92.45 & 95.31 & 64.09\\ 
		LEAM (linear) & 75.22 & 98.32  &91.75 & 93.43 & 61.03\\ 
		% Out Model Gated & 74.01 & - & -\\
		\hline
	\end{tabular}
	\vspace{-2mm}
	\caption{Test Accuracy on document classification tasks, in percentage. $^{\diamond}$ We ran Bi-BloSAN using the authors' implementation; all other results are directly cited from the respective papers.}
	\label{tab:accuracy}
	\vspace{-0mm}
\end{table*}

We compare with a variety of methods, including 
$(\RN{1})$ the bag-of-words in~\citep{zhang2015character}; 
$(\RN{2})$ sophisticated deep CNN/RNN models: large/small word CNN,  LSTM reported in~\citep{zhang2015character,dai2015semi} and deep CNN (29 layer) \citep{conneau2017very};
$(\RN{3})$ simple compositional methods: fastText \citep{joulin2016bag} and simple word embedding models (SWEM) \citep{shen2018on};
$(\RN{4})$ deep attention models: hierarchical attention network (HAN) \citep{yang2016hierarchical};
$(\RN{5})$ simple attention models: bi-directional block self-attention network
(Bi-BloSAN)~\citep{shen2018bi}.
The results are shown in Table~\ref{tab:accuracy}. 

\paragraph{Testing accuracy} Simple compositional methods indeed achieve comparable performance as the sophisticated deep CNN/RNN models. On the other hand, deep hierarchical attention model can improve the pure CNN/RNN models. The recently proposed self-attention network generally yield higher accuracy than previous methods. All approaches are better than traditional bag-of-words method. Our proposed LEAM outperforms the state-of-the-art methods on two largest datasets, \ie Yahoo and DBPedia. On other datasets, LEAM ranks the 2nd or 3rd best, which are similar to top 1 method in term of the accuracy. This is probably due to two reasons: 
$(\RN{1})$ the number of classes on these datasets is smaller, and
$(\RN{2})$ there is no explicit corresponding word embedding available for the label embedding initialization during learning. The potential of label embedding may not be fully exploited. As the ablation study, we replace the nonlinear compatibility~\eqref{eq:label_word_nonlinear} to the linear one in~\eqref{eq:label_word_linear} . The degraded performance demonstrates the necessity of spatial dependency and nonlinearity in constructing the attentions.

\begin{figure*}[t!] \centering
	\vspace{-0mm}
	\begin{tabular}{ccc}		
		\includegraphics[width=5.0cm]{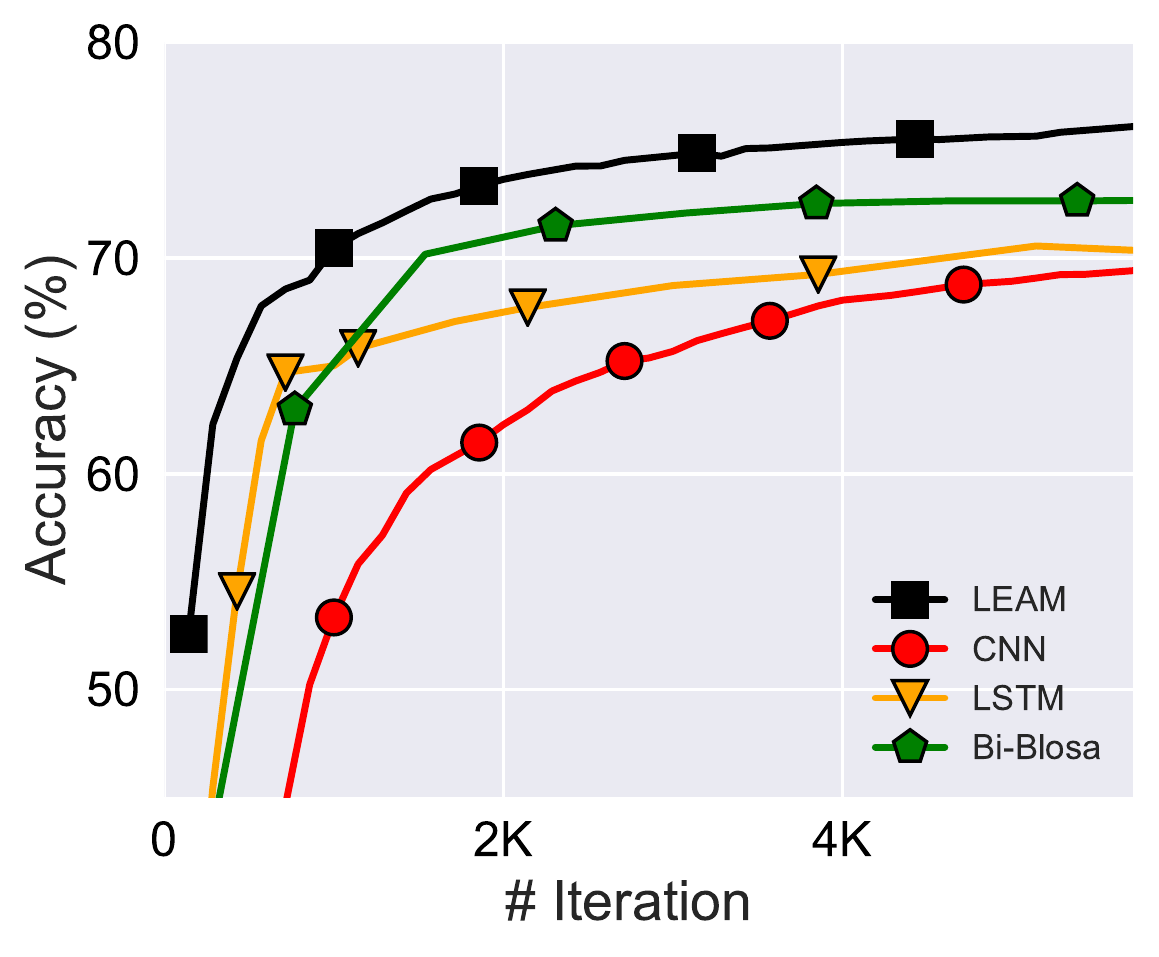} & \hspace{-0mm}
		\includegraphics[width=5.00cm]{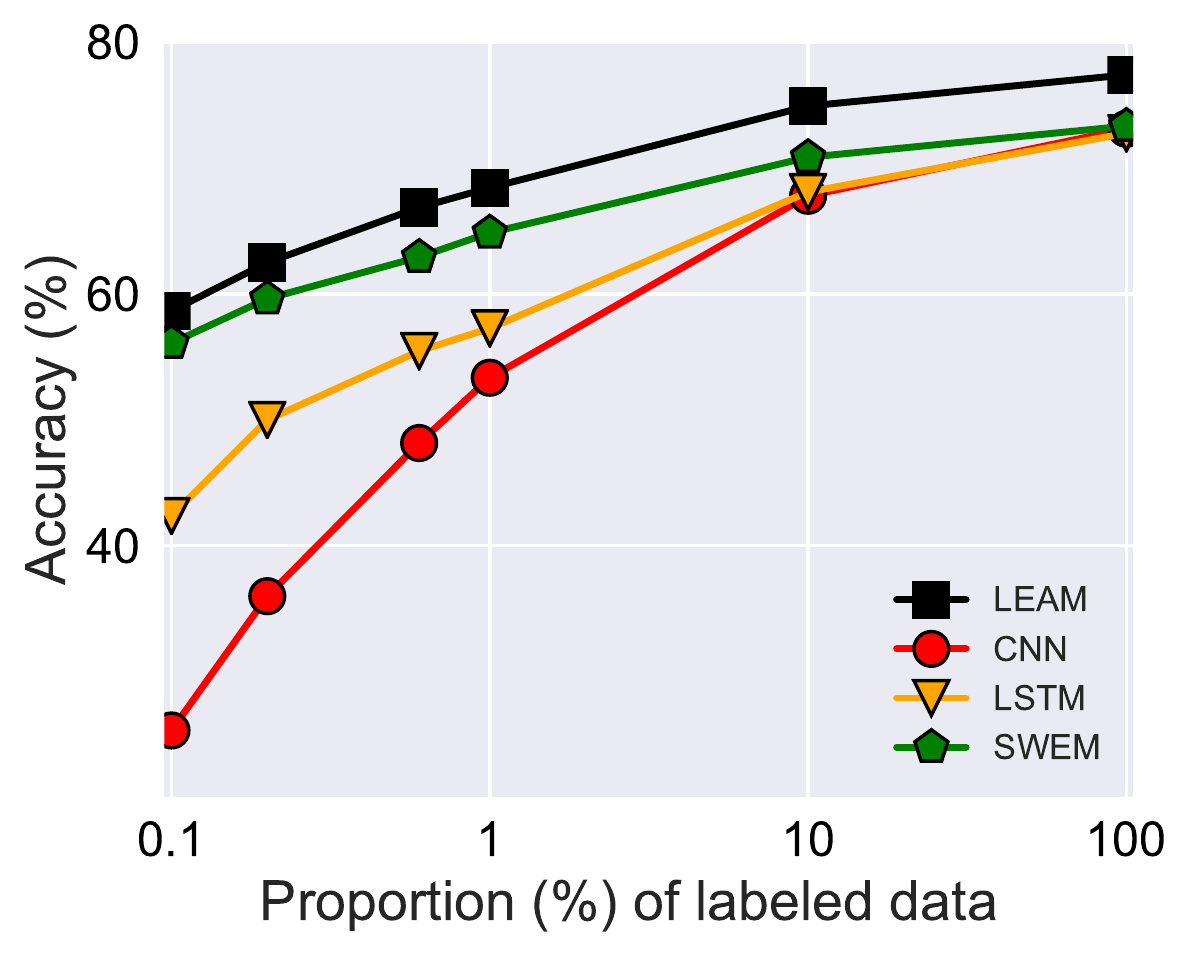} & 
		\includegraphics[width=5.00cm]{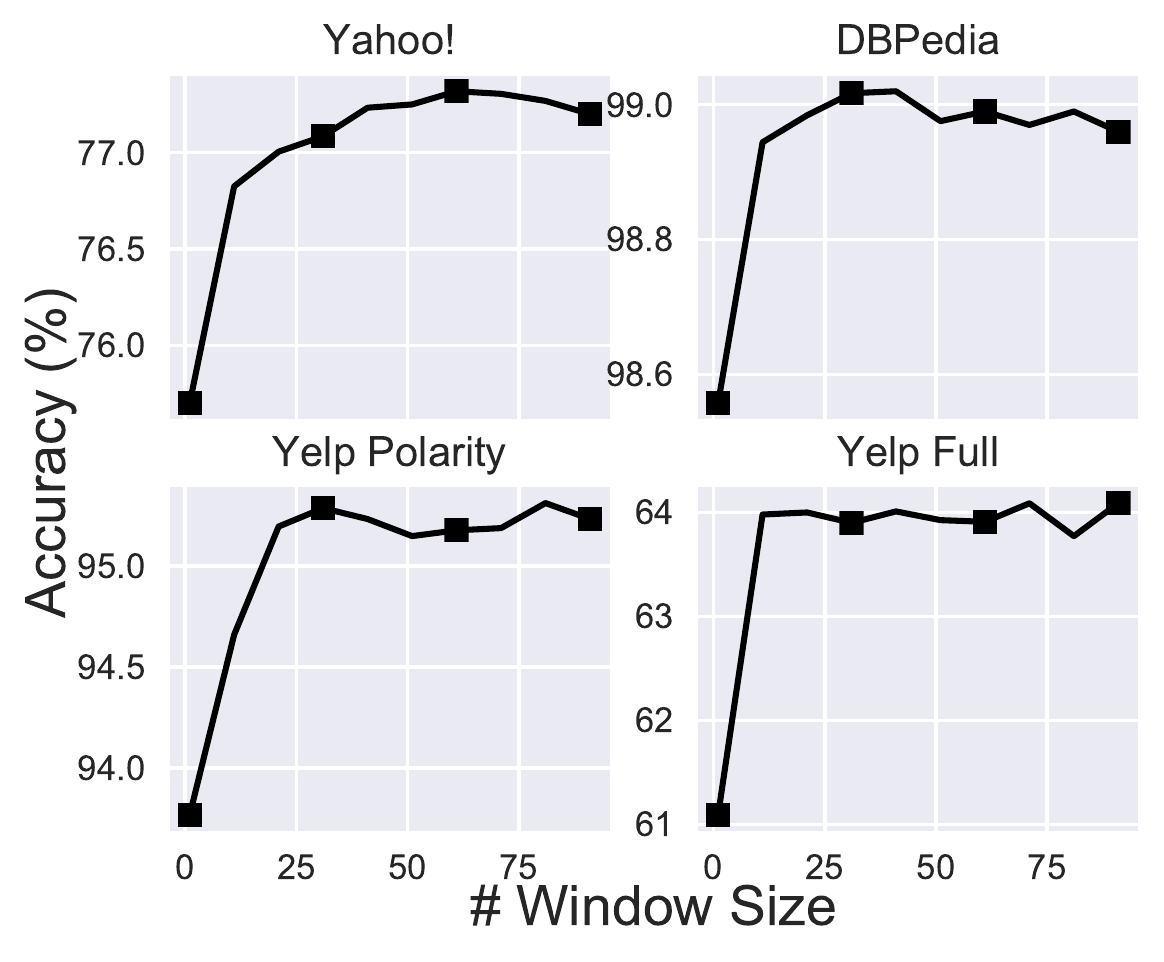} 
		\\
		\hspace{-0mm}
		(a) Convergence speed \vspace{0mm}  & 
		\hspace{-0mm}
		(b) Partially labeled data \hspace{-0mm}& 
		(c) Effects of window size \hspace{-0mm}
	\end{tabular}
	\vspace{-2mm}
	\caption{Comprehensive study of LEAM, including convergence speed, performance vs proportion of labeled data, and impact of hyper-parameter}
	\vspace{-0mm}
	\label{fig:leam_details}
\end{figure*}

% \paragraph{Speed and computational cost}
Nevertheless, we argue LEAM is favorable for text classification, by comparing the model size and time cost Table~\ref{tab:model_complexity}, as well as convergence speed in Figure~\ref{fig:leam_details}(a). The time cost is reported as the wall-clock time for 1000 iterations. LEAM maintains the simplicity and low cost of SWEM, compared with other models. LEAM uses much less model parameters, and converges significantly faster than Bi-BloSAN.
We also compare the performance when only a partial dataset is labeled, the results are shown in Figure~\ref{fig:leam_details}(b). LEAM consistently  outperforms other methods with different proportion of labeled data.

\begin{table}
	\centering
	\begin{adjustbox}{scale=0.9,tabular=l|cc,center}
		\hline
		\textbf{Model} & \textbf{\# Parameters} & \textbf{Time cost (s)}  \\
		\hline
		CNN & 541k & 171 \\
		LSTM & 1.8M & 598  \\
		SWEM & 61K  & 63  \\
		Bi-BloSAN & 3.6M &  292  \\ \hline
		LEAM & 65K &  65 \\ 
		\hline
	\end{adjustbox}
	\caption{Comparison of model size and speed.}
	\vspace{-4mm}
	\label{tab:model_complexity}
\end{table}

\paragraph{Hyper-parameter} Our method has an additional hyperparameter, the window size $r$ to define the length of ``phase'' to construct the attention. Larger $r$ captures long term dependency, while smaller $r$ enforces the local dependency. We study its impact in Figure~\ref{fig:leam_details}(c). The topic classification tasks generally requires a larger $r$, while sentiment classification tasks allows relatively smaller $r$. One may safely choose $r$ around $50$ if not finetuning. We report the optimal results in Table~\ref{tab:accuracy}.  

\begin{figure*}[t!] \centering
	\vspace{-0mm}
	\begin{tabular}{ccc}		
		\includegraphics[width=5.50cm]{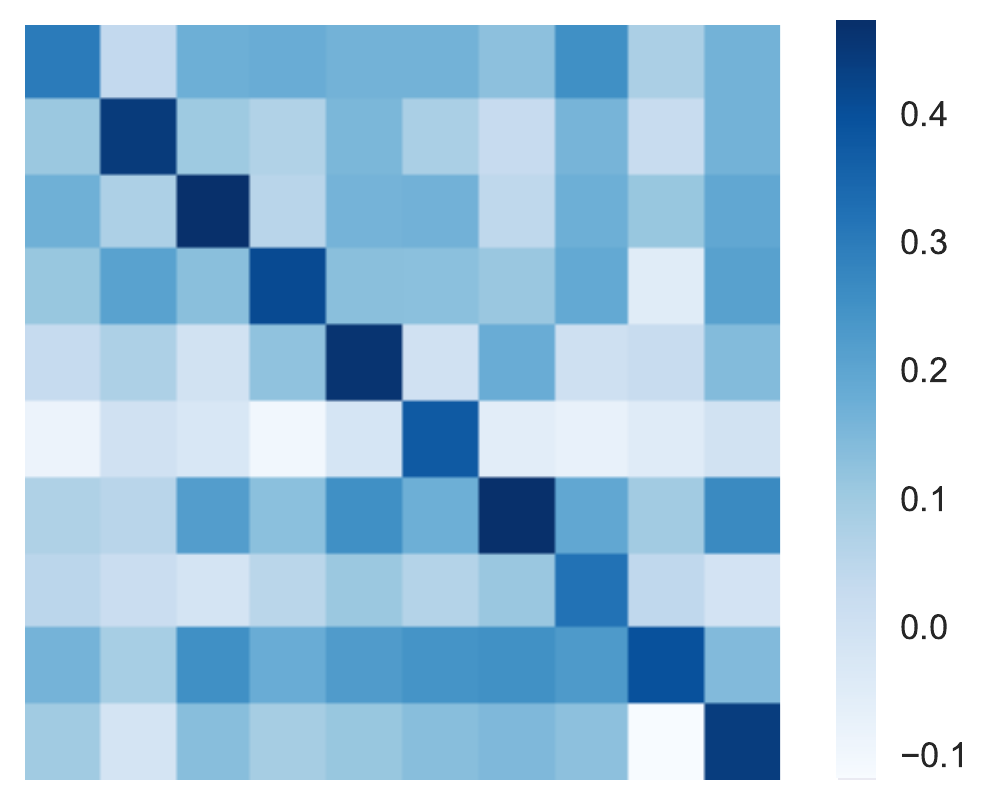} & \hspace{-5mm}
		\includegraphics[width=6.30cm]{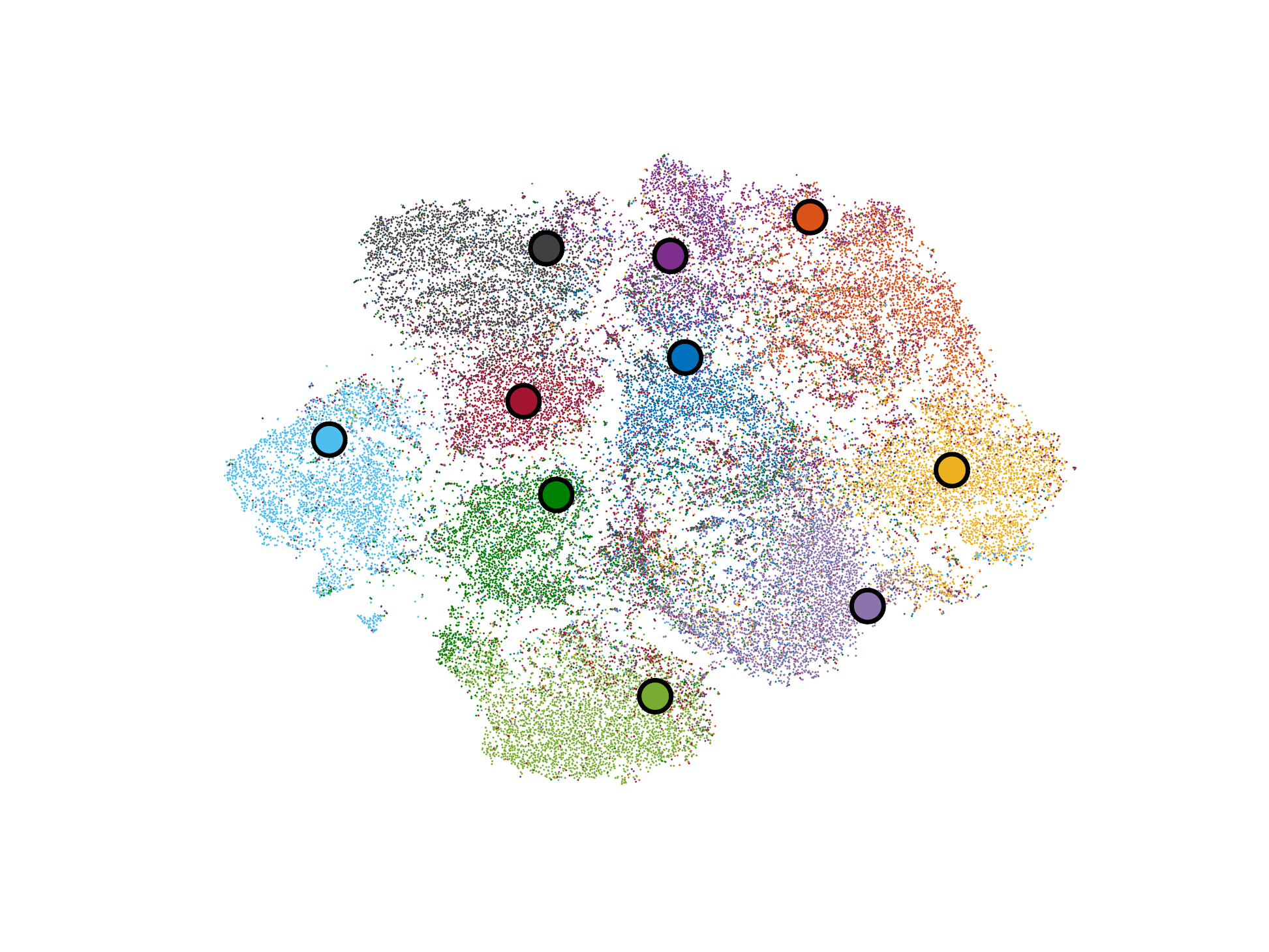} 
		\includegraphics[width=3.50cm]{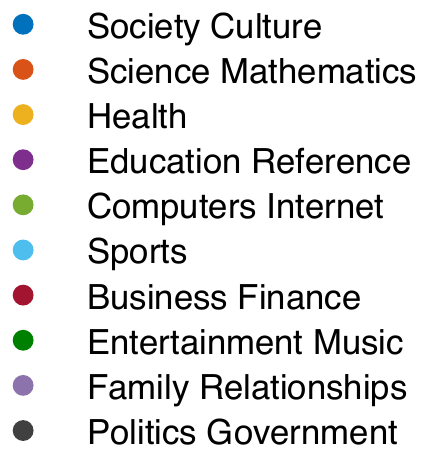} 
		\\
		\hspace{-5mm}
		(a) Cosine similarity matrix \vspace{0mm}  & 
		\hspace{-15mm}
		(b) t-SNE plot of joint embeddings \hspace{-0mm}& 
	\end{tabular}
	\vspace{-2mm}
	\caption{Correlation between the learned text sequence representation $\zv$ and label embedding $\Vmat$. (a) Cosine similarity matrix between averaged $\bar{\zv}$ per class  and label embedding $\Vmat$, and (b) t-SNE plot of joint embedding of text $\zv$ and labels $\Vmat$. }
	\vspace{-2mm}
	\label{fig:text_label_correlation}
\end{figure*}

\subsection{Representational Ability}
\paragraph{Label embeddings are highly meaningful} 
To provide insight into the meaningfulness of the learned representations, in Figure~\ref{fig:text_label_correlation} we  visualize the correlation between label embeddings and document embeddings based on the Yahoo dateset. First, we compute the averaged document embeddings per class: 
$\bar{\zv}_k =  \frac{1}{ |\Scal_k| }  \sum_{i \in \Scal_k}\zv_i$, 
where $\Scal_k$ is the set of sample indices  belonging to class $k$. Intuitively, $\bar{\zv}_k$ represents the center of embedded text manifold for class $k$. Ideally, the perfect label embedding $\cv_k$ should be the representative anchor point for class $k$. We compute the cosine similarity between $\bar{\zv_k}$ and $\cv_k$ across all the classes, shown in Figure~\ref{fig:text_label_correlation}(a). The rows are averaged per-class document embeddings, while columns are label embeddings. Therefore, the on-diagonal elements measure how representative the learned label embeddings are to describe its own classes, while off-diagonal elements reflect how distinctive the label embeddings are to be separated from other classes. The high on-diagonal elements and low off-diagonal elements in Figure~\ref{fig:text_label_correlation}(a) indicate the superb ability of the label representations learned from LEAM.

Further, since both the document and label embeddings live in the same high-dimensional space, we use t-SNE~\citep{maaten2008visualizing} to visualize them on a 2D map in Figure~\ref{fig:text_label_correlation}(b). Each color represents a different class, the point clouds are document embeddings, and the label embeddings are the large dots with black circles. As can be seen, each label embedding falls into the internal region of the respective manifold, which again demonstrate the strong representative power of label embeddings.

\paragraph{Interpretability of attention} Our attention score $\betav$ can be used to highlight the most informative words \wrt the downstream prediction task. We visualize two examples in Figure \ref{fig:attention}(a) for the Yahoo dataset. The darker yellow means more important words. The 1st text sequence is on the topic of ``Sports'', and the 2nd text sequence is ``Entertainment''. The attention score can correctly detect the key words with proper scores.

\begin{table*}[t!]
	\centering
	\begin{adjustbox}{scale=1.00,tabular=l|cc|cc|c,center}
		\hline
		& \multicolumn{2}{|c|}{AUC}  & \multicolumn{2}{|c|}{F1} & \\
		Model &  Macro& Micro &  Macro & Micro &\ P@5   \\ \hline
		Logistic Regression  & 0.829 & 0.864 & 0.477 & 0.533 &0.546 \\		
		Bi-GRU  &  0.828 & 0.868 &0.484 & 0.549 & 0.591 \\		
		CNN~\cite{kim2014convolutional} & 0.876 & 0.907 &  0.576  &   0.625 &  0.620 \\
		C-MemNN~\cite{prakash2017condensed} & 0.833 & - & - &  - & 0.42 \\
		Attentive LSTM~\cite{shi2017towards} & - &  0.900 & - &  0.532 & - \\
		CAML~\cite{mullenbach2018explainable} & 0.875 & 0.909  & 0.532  & 0.614 & 0.609 \\ \hline
		LEAM &  {\bf 0.881}  &  {\bf 0.912} &   0.540  &  0.619   &  0.612  \\
		\hline
	\end{adjustbox}
	\caption{Quantitative results for doctor-notes multi-label classification task.}
	\label{tab:multilabel-results}
	\vspace{-3mm}
\end{table*}

\subsection{Applications to Clinical Text}
To demonstrate the practical value of label embeddings,
we apply LEAM for a real health care scenario: medical code prediction on the Electronic Health Records dataset. A given patient may have multiple diagnoses, and thus multi-label learning is required. 

Specifically, we consider an open-access dataset, $\mathtt{MIMIC}$-$\mathtt{III}$~\cite{johnson2016mimic}, which contains text and structured records from a hospital intensive care unit. 
Each record includes a variety of narrative notes describing a patient’s stay, including diagnoses and procedures. 
They are accompanied by a set of metadata codes from the International Classification of Diseases (ICD), which present a standardized way of indicating
diagnoses/procedures.
To compare with previous work, we follow~\cite{shi2017towards,mullenbach2018explainable}, and preprocess a dataset consisting of the most common 50 labels. It results in 8,067 documents for training, 1,574 for validation, and 1,730 for testing.

\paragraph{Results}

We compare against the three baselines: a logistic regression model with bag-of-words, a bidirectional gated recurrent unit (Bi-GRU) and a single-layer 1D convolutional network~\cite{kim2014convolutional}. We also compare with three recent methods for multi-label classification of clinical text, including Condensed Memory Networks (C-MemNN)~\cite{prakash2017condensed}, Attentive LSTM~\cite{shi2017towards} and Convolutional Attention (CAML)~\cite{mullenbach2018explainable}.

To quantify the prediction performance, we follow~\cite{mullenbach2018explainable} to consider the micro-averaged and macro-averaged F1 and area under the ROC curve (AUC), as well as the precision at $n$ ({\it P@$n$}). Micro-averaged values are calculated by treating each (text, code) pair as a separate prediction. Macro-averaged values are calculated by averaging metrics computed per-label. P@$n$ is the fraction of the $n$ highestscored labels that are present in the ground truth.

The results are shown in~Table \ref{tab:multilabel-results}. LEAM provides the best AUC score, and better F1 and P@5 values than all methods except CNN. CNN consistently outperforms the basic Bi-GRU architecture, and the logistic regression baseline performs worse than all deep learning architectures.

\begin{figure}[t!] \centering
	\vspace{-0mm}
	\begin{tabular}{c}		
		\hspace{-5mm}
		\includegraphics[width=8.40cm]{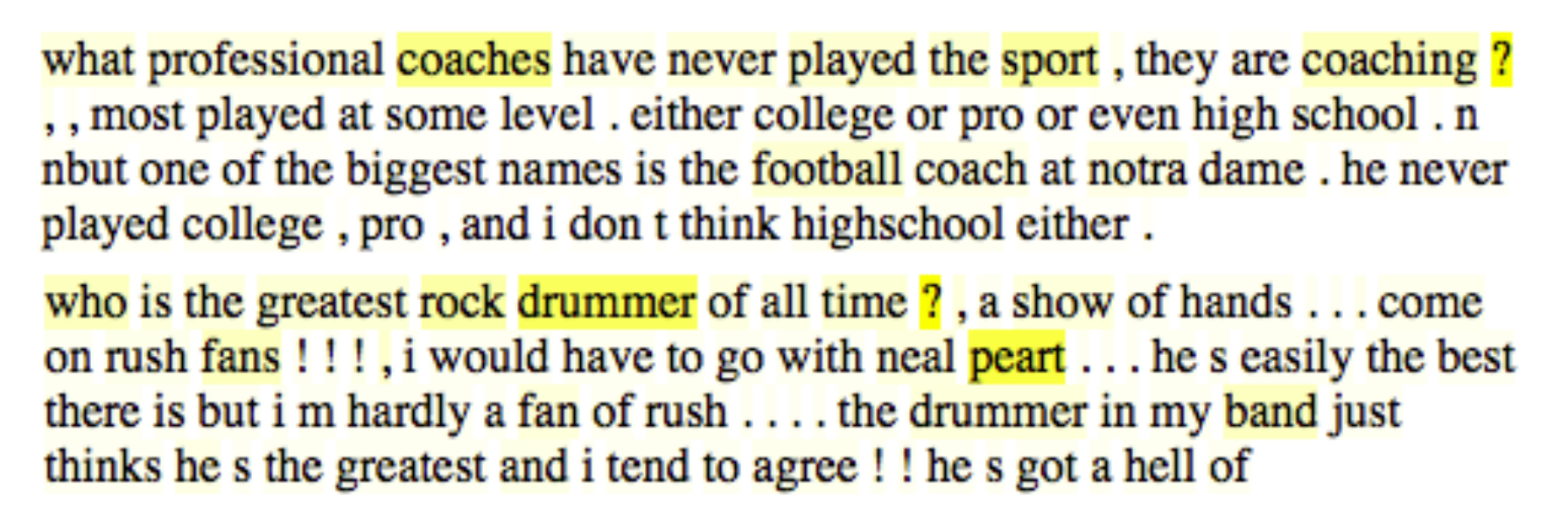} \\
		(a) Yahoo dataset  \vspace{0mm}  \\
		\hspace{-5mm}
		\includegraphics[width=8.40cm]{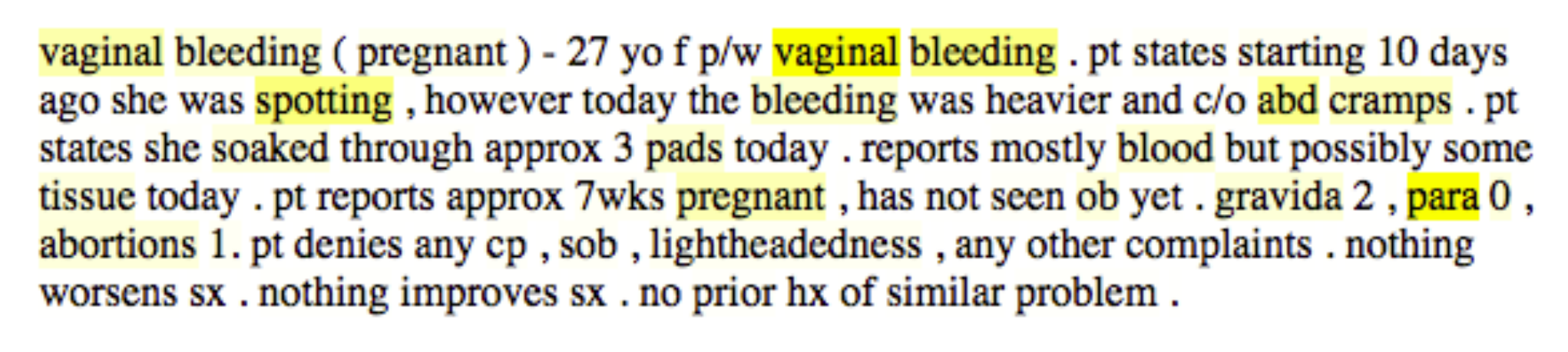}   \\
		(b) Clinical text  \hspace{-0mm}
	\end{tabular}
	\vspace{-1mm}
	\caption{Visualization of learned attention $\betav$.}
	\vspace{-0mm}
	\label{fig:attention}
\end{figure}

We emphasize that the learned attention can be very useful to reduce a doctor's reading burden. As shown in Figure \ref{fig:attention}(b), the health related words are highlighted.

\section{Conclusions}
\label{sec:conclusions}
In this work, we first investigate label embeddings for text representations, 
and propose the label-embedding attentive models. It embeds the words and labels in the same joint space, and measures the compatibility of word-label pairs to attend the document representations.
The learning framework is tested on several large standard datasets and a real clinical text application. Compared with the previous methods, our LEAM algorithm requires much lower computational cost, and achieves better if not comparable performance relative to the state-of-the-art. The learned attention is highly interpretable: highlighting the most informative words in the text sequence for the downstream classification task.

\paragraph{Acknowledgments} This research was supported by DARPA, DOE, NIH, ONR and NSF.

% \section*{Acknowledgments}

\newpage
%\bibliography{acl2018}
\bibliography{acl2018}
\bibliographystyle{acl_natbib}

%\newpage
%\appendix
%
%\section{Supplemental Material}
%\label{sec:supplemental}
%
%\section{More Results}

\end{document}

%% file: mlVecMat.tex
\usepackage{url}

\usepackage{amsmath}
\usepackage{amssymb}
\usepackage{wrapfig}
\usepackage{bm}
\usepackage{subcaption}
\usepackage{psfrag}
\usepackage{epstopdf}
\usepackage{multirow}
\usepackage{mathtools} 

\usepackage{verbatim}

\usepackage{color}
\usepackage{tikz}
\usetikzlibrary{arrows,shapes,snakes,automata,backgrounds,fit,petri}
\usepackage{adjustbox}

\usepackage[lined,boxed,commentsnumbered,ruled]{algorithm2e}

\newcommand{\ie}[0]{\emph{i.e., }}

\newcommand{\eg}[0]{\emph{e.g., }}

\newcommand{\etc}[0]{\emph{etc. }}
\newcommand{\wrt}[0]{\emph{wrt }}

\newcommand{\beq}{\vspace{0mm}\begin{equation}}
\newcommand{\eeq}{\vspace{0mm}\end{equation}}
\newcommand{\beqs}{\vspace{0mm}\begin{eqnarray}}
\newcommand{\eeqs}{\vspace{0mm}\end{eqnarray}}
\newcommand{\barr}{\begin{array}}
\newcommand{\earr}{\end{array}}

\newcommand{\Cmat}{{\bf C}}

\newcommand{\Gmat}{{\bf G}}

\newcommand{\Vmat}[0]{{{\bf V}}}
\newcommand{\Wmat}[0]{{{\bf W}}}
\newcommand{\Xmat}[0]{{{\bf X}}}

\newcommand{\bv}[0]{{\boldsymbol{b}}}
\newcommand{\cv}[0]{{\boldsymbol{c}}}

\newcommand{\ev}[0]{{\boldsymbol{e}}\xspace}

\newcommand{\mv}[0]{{\boldsymbol{m}}}

\newcommand{\uv}{\boldsymbol{u}}
\newcommand{\vv}{\boldsymbol{v}}

\newcommand{\xv}{\boldsymbol{x}}
\newcommand{\yv}{\boldsymbol{y}}
\newcommand{\zv}{\boldsymbol{z}}

\newcommand{\betav}[0]{{\boldsymbol{\beta}}}

\newcommand{\thetav}{\boldsymbol{\theta}}

\newcommand{\R}{\mathbb{R}}

\newcommand{\Xcal}{\mathcal{X}}
\newcommand{\Ycal}{\mathcal{Y}}

\newcommand{\Scal}{\mathcal{S}}

\newcommand{\Fcal}{\mathcal{F}}

\ifx\assumption\undefined
\newtheorem{assumption}{Assumption}
\fi